# CEFER: A Four Facets Framework based on Context and Emotion embedded features for Implicit and Explicit Emotion Recognition

**Fereshteh Khoshnam[1], Ahmad Baraani-Dastjerdi[2,*], M.J. Liaghatdar[3]**

[1]Department of Software Engineering, Phd Candidate, University of Isfahan, Isfahan, Iran. E-mail address: f.khoshnam94@eng.ui.ac.ir

[2]Department of Software Engineering, Dr, University of Isfahan, Hezar-Jerib Ave., Isfahan, 81746-73441, Iran. E-mail address: ahmadb@eng.ui.ac.ir

[3]Faculty of Education and Psychology, Dr, University of Isfahan, Hezar-Jerib Ave., Isfahan, 81746-73441, Iran.

## Abstract

People's conduct and reactions are driven by their emotions. Online social media is becoming a great instrument for expressing emotions in written form. Paying attention to the context and the entire sentence help us to detect emotion from texts. However, this perspective inhibits us from noticing some emotional words or phrases in the text, particularly when the words express an emotion implicitly rather than explicitly. On the other hand, focusing only on the words and ignoring the context results in a distorted understanding of the sentence meaning and feeling. In this paper, we propose a framework that analyses text at both the sentence and word levels. We name it CEFER (Context and Emotion embedded Framework for Emotion Recognition). Our four approach facets are to extracting data by considering the entire sentence and each individual word simultaneously, as well as implicit and explicit emotions. The knowledge gained from these data not only mitigates the impact of flaws in the preceding approaches but also it strengthens the feature vector. We evaluate several feature spaces using BERT family and design the CEFER based on them. CEFER combines the emotional vector of each word, including explicit and implicit emotions, with the feature vector of each word based on context. CEFER performs better than the BERT family. The experimental results demonstrate that identifying implicit emotions are more challenging than detecting explicit emotions. CEFER, improves the accuracy of implicit emotion recognition. According to the results, CEFER perform 5% better than the BERT family in recognizing explicit emotions and 3% in implicit.

# 1. Introduction

People's lifestyles are changing due to the use of the Internet and various social media platforms. These changes have had an impact on practically the entire world daily routines. People use the internet to work, learn, shop, and socialize. It has also influence on how individuals communicate their opinions, and emotions; they are continuously posting their thoughts, feelings, and emotions on social media.

Emotions are the most important factor in human behavior. Emotion is a vital component of a person's reaction to both external and internal events. It causes him to reconsider his thoughts and perspectives [1, 2]. Therefore, analysis of publicly available data on platforms, which is typically in the form of document collections -often short texts- appears to be vital and plays a significant role in giving reliable public opinion insights. Researches in this field are valuable. Major domains of researches in this field are: health monitoring [3, 46, 47], election forecasting [4], financial markets [5], student surveys [6], and software product design [7, 50].

To recognize emotions from textual data, emotion analysis systems use a number of learning approaches: including lexical-based [8], machine / deep learning [9,10], a combination of lexical and machine learning [11], and concept-based learning approaches [12,13].

The task of creating systems that can automatically analyze natural language in order to understand emotional content is challenging and fraught with difficulties. A general problem is modeling human emotional behavior and recognizing the boundaries between different emotions in automated text-based systems [14].

Another key challenge is determining the text emotions through proper analysis. Two levels of analyses exist: sentence-level and word-level. The sentence-level is concerned with how to evaluate the context and applies the sentence meaning [4, 5, 6]. A word-level that necessitates concentrating on individual words which may or may not contain any emotions. These emotions may be expressed implicitly or explicitly.

Writers may avoid using explicit emotional words in the text in favor of using words that express the senses implicitly [15, 51]. Implicit emotions, according to researchers, influence the intensity and duration of emotional responses. In addition, the combination of implicit and explicit emotions allows complex emotional states to be deduced [48, 49]. As a result, it appears that broadening the scope of research to identify implicit emotions is required [51].

All of the aforementioned factors emphasize the necessity of recognizing both implicit and explicit emotions, as well as accurate text analysis at both the sentence and word levels. The current research is focused on text-based emotion mining. We propose a framework which we name CEFER that all four aforementioned approach facets are considered. Some key contributions of CEFER are as follows:

1-Constructing the feature vector of each word using the concept of the entire sentence and context. In this situation, the semantic of the entire sentence will be implicitly stored in the feature vector of each word.

2- Constructing the emotional vector of words based on implicit and explicit emotions at the same time

3- Studying the results of integrating emotional knowledge with extracted feature vectors.

4- Experimental results show that CEFER works well in both implicit and explicit emotion recognition.

5- CEFER can identify emotions in short English texts and classify them into Plutchik eight primary categories (Joy, Trust, Fear, Surprise, Sadness, Disgust, Anger, and Anticipation).

6- The emotional vector and the extracted feature vector for each word are used to construct the meta-embedding.

The evaluation of the proposed framework on three valid datasets shows that CEFER performs better than many previous works in detecting explicit and implicit emotions. CEFER, on the other hand, can simultaneously identify both types of emotion that none of the previous works have. Of course, except for DuFER, which is the previous work of the authors of the article. In addition, compared to the results obtained from transformers, CEFER shows a detection improvement of between 3% and 5%.

The organization of this article is as follows: First, we will review the related works. Subsequently, Section 3 provides more details on CEFER architecture. Experimental and evaluation results will be discussed in Section 4. Conclusion is presented in Section 5.

## 2. Related work

Emotions have been explored by psychologists in terms of definition and classification. Ekman [17], who recognized six primary emotions, and Plutchik [18], who classified emotions into eight broad groups, are two prominent and well-known definitions in this topic. Moreover, emotions are divided into two main types, explicit and implicit [36]. Many studies on text-based emotion recognition focus on explicit emotions. Some studies focus on implicit emotions, while only a few about both types of emotions. These studies used a variety of deep learning and machine learning methodologies.

Machine learning approaches have generally recognized explicit emotions [16, 19, 21, 22, 57], while a small number of research methods have identified implicit emotions. The two methods which identified implicit emotions are presented in the IEST competition, which came in 15th and 19th place among the 30 teams [20]. Finally, a few studies have been done, such as DuFER [15], to identify both implicit and explicit emotions at the same time.

In this study, the deep learning approach is applied. Therefore, studies that have employed deep learning methods are reviewed. Levels of text processing (sentence and word levels) and types of emotions (explicit and implicit) are explored.

Park et al. [23] concentrated on modeling emotional representations within the text at the sentence and word level. Neural network models were used as feature extractors. They used these representations for classic machine learning methods like support vector regression and logistic regression to transfer emotional knowledge. The emoticons and hashtags in their labeled corpuses were used. The problem is that their model only predicts emoji clusters with a maximum accuracy of 61.0 percent. As a result, the emotionally gathered knowledge is also less accurate only four of Plutchik eight main emotions (happy, sad, fear, and anger) are recognized. Furthermore, the work was focused on the expressive feelings of the characters. The explicit emotions of text are only considered.

Chatterjee et al. [24] suggested a deep learning approach for recognizing three emotions in textual dialogue: happy, sad, and angry. The sad emotion is the one that is most accurate (80.79 percent). This study is focused on recognizing explicit emotions. Implicit emotions are not considered. The method combines semantic and sentiment representations of words to better recognize explicit emotions. However, the limitation of model is that the context of conversation and the sentence features are not considered.

EmoDet2 [25] can recognize and classify the explicit emotions in English text conversation into four groups (happy, sad, angry, etc.). GloVe word embedding, BERT embedding, and a set of psycholinguistic features make up the main input of system. The suggested system combines the BiLSTM neural network with a fully connected neural network architecture. EmoDet2's results demonstrate that it has improved significantly (F1=0.748) over the base model submitted by the Semeval-2019 organizers (F1 = 0.58). However, semantic features are extracted by transforming the entire text into a single vector. EmoDet2 does not really involve the specific features of words and sentences. Furthermore, this approach has not recognized or even paid attention to the implicit emotions in order to find the explicit emotions.

Illendula and Sheth [26] studied the effect of emojis and images on the accuracy of categorizing explicit emotions in social media posts. They used the BiLSTM model with the attention mechanism. They fed it with extracted features from the images, FastText embedding and EmojiNet. They were able to reach a 72 percent accuracy rate. The explicit emotions (Anger, Fear, Joy, Love, Sad, Surprise, and Thankful) were recognized according to the features of the words. The number of words that caused the best accuracy was investigated but the sentence and context features were not taken into account. In addition, the implicit emotions of texts were not considered.

A classification approach based on deep neural networks, Bi-LSTM, CNN, and self-attention was proposed in [27]. For word encoding, three pre-trained word embeddings (GoogleEmb, GloVeEmb, and FastTextEmb) were employed and compared. The outcomes of different word embedding were compared. The four explicit emotions (anger, fear, joy, and sadness) were recognized with a maximum accuracy of 84 percent which were focused on the features of words. Context and sentence-level features were not taken into account in this study. It is also important to pay attention to implicit emotions. Low accuracy has resulted from a lack of attention to these factors.

Four types of sequence-based convolutional neural network models were proposed by Zahiri and Choi [28]. The sequence information encapsulated in conversation was used by these attentive networks. They also proposed detailed SCNN models that contained existing sequence information. The presented CNN is sequence-based and extracted features from the current speech are combined with previous ones. Sequence information was used to improve the classification of the seven emotions. This study attempted to maintain the sequence of conversation, however, this did not help paying attention to the concept at the sentence level. This leads not to pay attention to the implicit emotions in the text, resulting in a 54 percent accuracy rate.

BNet is a technique presented by Jabreel and Moreno [29] to turn the problem of emotion recognition into a binary classification problem. It has a 59 percent accuracy rate. Three embedding models and an attention function were used in this method. A set of key-value pairs was mapped to an output by the embedding module. BNet extracts features at the word level but sentence-level feature and the impact of implicit concepts throughout the sentence are not considered. The system recognizes eight emotions that it has aggressively encountered in relation to four of them. This means that the system has linked a small number of instances to these labels.

The application of natural language comprehension to emotion recognition tasks is investigated in [30]. Different transformers are fine-tuned by the writers (BERT, DistilBERT, RoBERTa, XLNet). They use a fine-grained emotion dataset for this purpose. They then assess each the performance of transformer appropriately. The major goal of this research is to look into the behavior of different transformer language models when it comes to recognize emotions. The authors have simply paid attention to the context in accordance with the purpose. The study ignores implicit emotions and concepts as well as word-level features. The findings demonstrate that these transformers perform similarly in the emotion recognition task while RoBERTa had the highest F1-score.

Schmidt et al. [31] examine the performance of different methods for classifying emotional texts in the context of German historical plays around 1800. Lexicon-based machine learning, FastText as static word embedding and various transformer models based on the BERT or ELECTRA architecture are among these methods. The recognition of 13 sub-emotions from six primary emotional categories is taken into account, resulting in a 66 percent accuracy rate. The entire text serves as the foundation for identifying emotion. Because many texts are relatively long, they contain a variety of emotions. So, the main reason of misdiagnosis in this study is mapping the entire text to only one emotion. As a result, shorter textual units such as sentences, phrases, and words must be prioritized. Furthermore, the writers' attention on the overall context has caused them to overlook the implicit emotions as well as the word-level and sentence-level analysis.

Krommyda et al. [52] identifies eight Plutchik emotions in short texts using emoji, keyword, and semantic analysis. The authors produce a dataset and use it to train classification models. They use an empirical investigation to determine the influential features and then created a model for recognizing emotions in social media posts. The authors' rationale for employing emotional and lexical dictionaries to extract word-level properties and semantic correlations are correct, resulting in a 91.9 percent accuracy. The sentence-level features and context, on the other hand, have received no attention.

The goal of [53] is to use deep learning and natural language processing to identify, examine, and understand the emotions expressed globally in the early months of the Covid. Model analysis and text-based emotional data analysis are both used in this research. The accuracy of model in recognizing emotions in Twitter texts is 80.33 percent. This categorization is based on 13 different emotions (anger, boredom, empty, enthusiasm, fun, happiness, hate, love, neutral, relief, sadness, surprise and worry). The lack of a distinct emotional model has resulted in the categorization of emotions into numerous groupings, lowering accuracy. Some of these emotions might be combined. Another issue is not paying attention to the implicit concepts, especially due to the limitations of the collected hashtags. The authors were able to discover emotions more accurately by addressing this issue. The essential point is that they may get better results if they focus on both word-level and sentence-level characteristics at the same time.

## 3. Methodology

Emotion recognition is modeled as a text classification problem that can assign one or more emotional labels to a sentence. In CEFER, the general idea for recognition emotions from short English texts is to use the meaning of the entire sentence and context as well as to pay attention to each word in particular. Thus, a proper and accurate label can be assigned to the text by simultaneously considering the implicit and explicit emotions in the text.

We used transformer-based word embedding in this study with considering four aspects: extracting data from the entire sentence and each individual word simultaneously as well as implicit and explicit emotions. The extraction of contextual information is required for the discovery of implicit emotions as well as the meaning of the sentence. Hence, transformer is a suitable technique to handle this issue [51]. Since transformers employ learning algorithms to increase classification accuracy, they also aid with more accurate classification of words and explicit emotions in the text [55]. The CEFER framework which is designed to recognize emotions in short texts, is explained in the following sections.

### 3.1. Overview of the framework

The main purpose of this study is to create a framework for inferring emotions from text and classifying texts based on their emotional content. The framework CEFER can recognize the correct label, even if there are no clear emotional words in the text. Figure 1 shows all part of the proposed framework CEFER: preprocessing tweets, extracting word features, inferring emotions using an emotional dictionary, building implicit or explicit emotion vectors of words, constructing meta-embedding, and recognizing the emotions of tweets. Detail of each parts is explained in the following.

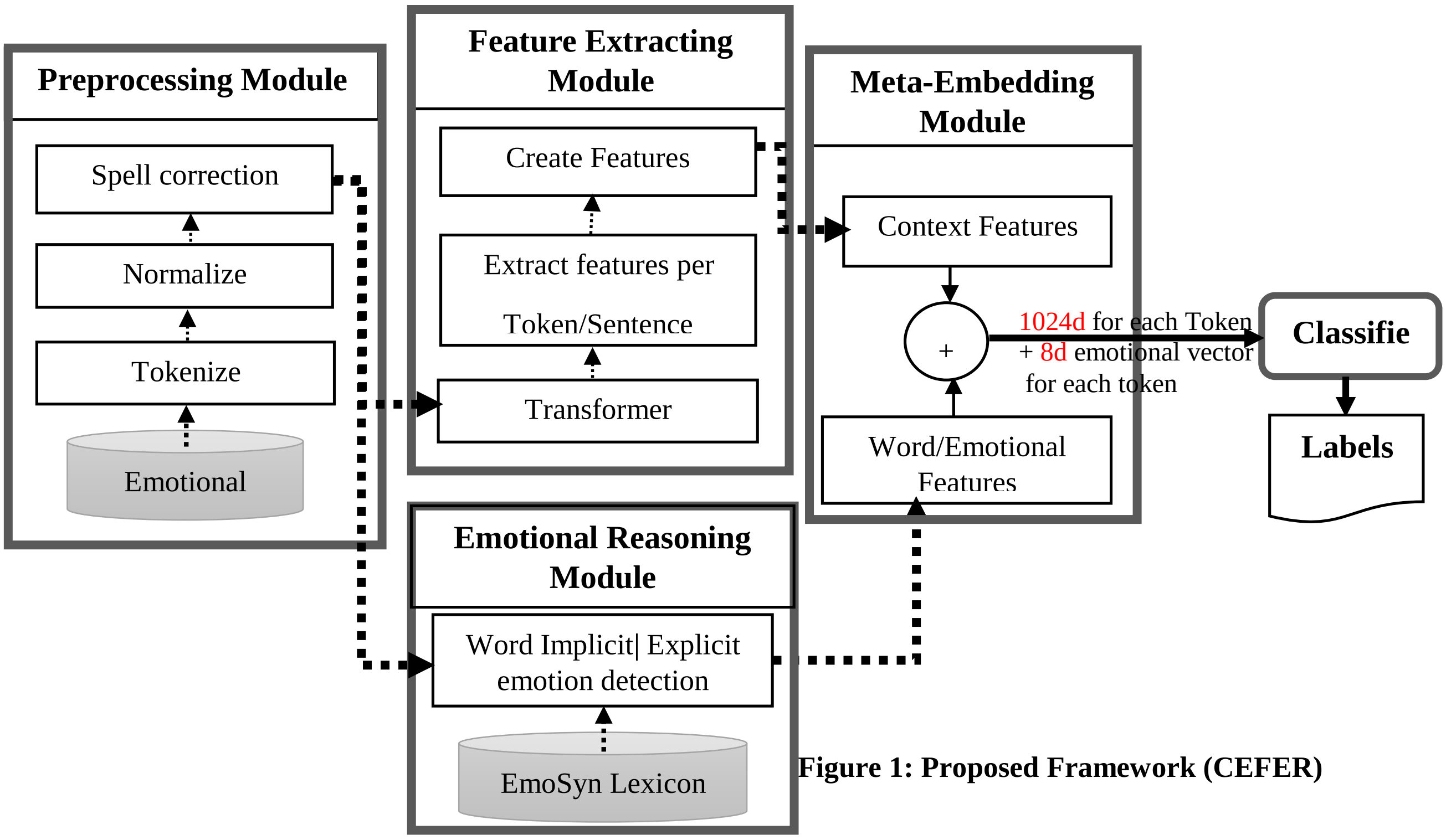

**Figure 1: Proposed Framework (CEFER)**

## 3.2. Preprocessing Module

Presence of hyperlinks and emojis in tweets are increase their meaning and mention. Also, they are used for questions and referring specific people. However, they are not necessary for recognizing emotions and they should be removed from texts of dataset. In order to do this, preprocessing of tweets is done. To begin preprocessing, any tokens beyond the main part of the tweet are deleted and then the tweets are normalized.

Another problem is character repetition in a word as well as misspelling words. To address this problem, each tweet is tokenized and the correctness of each token is checked against a dictionary. If a token is recognized invalid character repetition elimination and hashtag checking are done. For character repetition elimination, the Ekphrasis tool [54] is used. Characters that appear more than once are gradually eliminated to have a correct token. Hashtag checking are explained in Section 3.4.2 due to the use of several interconnected tokens and their specific writing form.

## 3.3. Feature Extraction Module

As said before, the goal of this study is to use word and sentence features and contextual concepts at the same time in order to obtain appropriate features for emotion recognition. In order to reach this goal, Feature Extraction module has three phases: transformer, feature extraction in sentences, and feature extraction in words. Each phase is explained in the following subsections.

### 3.3.1. **Transformer**

To extract word and sentence features, we use transformer in the CEFER because BERT [32] is the most explored transformer-based model for recognizing text-based emotions and a number of variants of that are proposed [27, 55]. Therefore, in experimental, we use BERT [32] and RoBERTa [33] transformers in both base and large modes in the CEFER. Both transformers have 12 hidden layers with 768 dimensions in base mode and 24 hidden layers with 1024 dimensions in large mode. The maximum input text length is considered to be 512.

RoBERTa and BERT learn different semantic information in different layers. In general, the shallower layers are learned more semantic information at the word level. The deeper layers are learned more generalized semantic information [34]. The BERT encoder produces a sequence of hidden states. For classification tasks, this sequence ultimately needs to be reduced to a single vector. To make this pooling scheme work, BERT prepends a [CLS] token (short for "classification") to the start of each sentence. It represents sentence-level features. The [CLS] token in the last layer is commonly assumed to contain semantic information to classify the entire sentence, ignoring a lot of information about the sentence. In CEFER, feature extraction is done in two methods: Sentence feature extraction using [CLS] and Word feature extraction.

### 3.3.2. Feature Extraction in Sentences

Both base and large models of BERT and RoBERTa transformers are used the entire sentence as input. The [CLS] output is used as a feature vector the sentence. This process is carried out in two steps.

- In the first step, data are sent into the transformer, and the [CLS] of the last layer is used as the feature vector of each tweet.
- In the second step, the [CLS] from the upper layers of transformer are received. These values are then combined in different ways in order to produce the best feasible combination feature vector.

The main goal of this phase is to feed the full sentence into the transformer as input, then analyze the features that the transformer returns for the entire sentence. The effectiveness of this method is then compared to that of other feature extraction methods. The accuracy of these feature vectors is shown in Section 4.

### 3.3.3. Feature Extraction in Words

The feature vector extraction of all the words in the text is performed at this phase in order to pay more attention to the words. By extracting the features of each word according to the sentence, it is feasible to have both context knowledge and semantic information about the words in the extracted features at the same time. This task is done in two steps.

- In the first step, the feature vector of each word is obtained from last layer of the transformer. Then, the entire sentence vector is created by combining word vectors.
- In the second step, the feature vectors of each word are obtained from the last four upper layers of the transformer which is similar to Section 3.3.2. After extracting all the feature vectors for each word, they are merged.

The task of recognizing emotions is then completed. Tables 3 and 4 in Section 4 demonstrate the results of emotion recognition accuracy using these feature vectors.

## 3.4. Emotional Reasoning Module

The usage of BERT variations demonstrates that language knowledge is acquired rapidly and consistently across most domains although facts are more domain-specific. Reasoning abilities are also not gained in a consistent manner. Many abilities are not learnt through RoBERTa's pre-training [35]. To extract the implicit emotions from the text, reasoning is required [36]. Inference can be added to the embedding methods since they need to be reinforced in reasoning. As a result, in the CEFER, identifying words with implicit emotions at different levels has helped in reaching this inference. To do so, we use EmoSyn Emotional Dictionary which is explained in the following subsections.

### 3.4.1. EmoSyn Emotional Dictionary

The EmoSyn is created as an emotional dictionary with the aim of extracting implicit and explicit emotions from words. It is made from three emotional dictionaries: WordNet (hierarchical model), NRC Emotion Lexicon (based on Plutchik, which is a categorical model) and NRC Hashtag Emotion Lexicon [15]. EmoSyn, actually combines hierarchical and categorical emotional models.

Based on the Plutchik model, this dictionary has eight categories of words and emotional phrases (Anger, Fear, Joy, Sadness, Anticipation, Trust, Surprise, and Disgust). Each category contains words and phrases that can convey a feeling either implicitly or explicitly. EmoSyn is made up of emotional words that represent each of the eight categories of Plutchik. Synonyms for these words are taken from three mentioned dictionaries at three levels, and added to each category. The reason for exploring up to three levels is that even if words do not represent an emotion explicitly, they can convey it implicitly. That is, they may have emotional synonyms in the first, second, or third level, so words can implicitly express an emotion.

### 3.4.2. Creating Emotion Vector using EmoSyn

In the following, We show how an emotional vector is provided for each word. The structure of emotion vector for each token or word is shown in Figure 2. The emotional vector of each token has eight elements because the Plutchik emotional model which contains eight basic emotions using in this study.

8d

| | Token1 | | | | | | | | Token2 | ......... |
|---|---|---|---|---|---|---|---|---|---|---|
| | E1 | E2 | E3 | E4 | E5 | E6 | E7 | E8 | | |
| Tweet1 | | | | | | | | | | |
| Tweet2 | . . | | | | | | | | | |

**Figure 2: Emotional Vector**

If x is an inferred word, the emotional vector of the word x is provided as follow. First, word x will be searched throughout all EmoSyn categories. If x belongs to two categories y and z, it means that x has two emotions y and z. Then only for the values associated to these two emotions, y and z, '1' is added in the emotional vector while the rest of the values of the emotional vector becomes '0'. If the token x is a hashtag, this is exactly the way which we use to provide its emotional vector. Since in EmoSyn for this form of emotional data, the NRC Hashtag Emotion Lexicon vocabulary is used. For example, if we have Example 1 and we use EmoSyn to generate an emotional vector for each token in Example1. The emotional vectors will look like what are shown in Figure 3.

*Example1: I don't want to sit here any longer.*

Figure 3 depicts two samples of emotional vectors of tokens of the Example 1. Four emotions- anger, fear, anticipation and disgust- are obtained for token "any", while five emotions-anger, fear, disgust, sadness, and surprise- are gained for token “want”. These vectors are then employed as emotional token vectors to assist in sentence recognition.

The emotions in Example 1 are anger and disgust according to [36]. However, the sentence is mistakenly labeled as neutral using the proposed approach for recognizing implicit emotions in [36]. While CEFER predicts the anger emotion for Example 1 which is correct.

| | **Want** | | | | | | | | **Any** | | | | | | | | . . . |
|---|---|---|---|---|---|---|---|---|---|---|---|---|---|---|---|---|---|
| | **Anger** | **Fear** | **Anticipation** | **Disgust** | **Sadness** | **Joy** | **Trust** | **Surprise** | **Anger** | **Fear** | **Anticipation** | **Disgust** | **Sadness** | **Joy** | **Trust** | **Surprise** | . . . |
| **Sentence** | **1** | **1** | **0** | **1** | **1** | **0** | **0** | **1** | **1** | **1** | **1** | **1** | **0** | **0** | **0** | **0** | . . . |

**Figure 3: An example of extracting the emotional vector of tokens**

## 3.5. Meta-Embedding Module

Meta-embedding integrates different word embeddings and successfully improves system performance over equivalent models that use only one type of embedding [37]. Multiple embedding can improve performance in three ways:

1. Expands the model and the capacity of network.

2. Increases vocabulary size. That is, instead of using the initial randomization, the model utilizes a better beginning point.

3. The second embedding presents a different perspective on the word. The model is able to take use of diverse features based on differences in how each word is shown.

Several approaches for creating meta-embedding have been presented in [37, 38, and 39]. Despite the variety of methods that have been presented for meta-embedding, concatenation remains a very powerful method for meta-embedding [37]. It keeps the entire structure of the primary embeddings. Also, many new features make the concatenation results desirable. However, the dimensions are increased linearly where it can sometimes be problematic.

As stated in Section 3.4.2, in providing an emotional vector, an 8-dimensional vector, $W_{Emotional}$, is generated for each word that expresses the emotions encoded in that word. The emotional 8-digit vector $W_{Emotional}$ is concatenated to the feature vector acquired from RoBERTa, $W_{RoBERTa}$, to create the feature vector of the proposed approach (Equation 1). This gives CEFER a particular emotional meta-embedding.

$$Meta\ (W) = Embedding\ (W_{RoBERTa}) + Embedding\ (W_{Emotional}) \qquad \text{Equation (1)}$$

This vector is used to improve the recognition of emotions in short texts.

## 4. Experimental Results

The three standard datasets for evaluation of the framework, CEFER, are introduced in Section 4.1. Emotional tweets in the first and second datasets contain words that express the writer's feelings explicitly. The final dataset contains tweets that have been stripped of explicit emotional words. The tweets in this dataset, according to its creators, implicitly express the author's feelings.

### 4.1. Datasets

Three different datasets are used to evaluate the proposed framework.

1. The Emotion Intensity Dataset [40] categorizes its tweets into four emotional classes (joy, sadness, fear, and anger). Each tweet contains an emotional label and a numerical value that represents the intensity of the emotion (Table 1).

*Table 1: The number of tweets*

| ***Dataset*** | **Train** | **Dev** | **Test** | **Total** |
|---|---|---|---|---|
| *Emotion Intensity dataset* | 3613 | 342 | 3142 | 7097 |

2. Hasan collected the EmoTex dataset [16] from Twitter. Tweets were automatically categorized based on the Circumplex emotional model depending on the emotions of their writers. Table 2 displays the dataset's attributes.

*Table 2: Number of tweets collected as labeled data after preprocessing*

| ***Class*** | **Happy-Active** | **Happy-Inactive** | **Unhappy-Active** | **Unhappy-Inactive** | **Total** |
|---|---|---|---|---|---|
| ***EmoTex dataset*** | 34000 | 30000 | 37000 | 34000 | 135000 |

3. The third dataset has been developed in preparation for the WASSA 2018 IEST [41] (Table 3). For the first time, the proposed task requires systems to predict emotions of tweets without explicit access to words expressing emotions. The Implicit Emotion Shared Task (IEST) is named after this purpose since the systems must infer the emotion mostly from the context. Every tweet contains an explicit emotive word that has been masked. The training data for this competition is only accessible to participants who have usernames and passwords. In this study, we only use its test data for both training and testing, which included 28757 labelled tweets.

*Table 3: Distribution of IEST data*

| ***Dataset*** | **Train** | **Trial** | **Test** | **Total** |
|---|---|---|---|---|
| ***IEST dataset*** | 153383 | 9591 | 28757 | 191731 |

## 4.2. Results

The results of experiments which are done with the CEFER are presented in the following.

### 4.2.1. Results of Feature Extraction in Sentences

As mentioned in Section 3.3.1, the data are first sent into the transformer, and the [CLS] output of the final layer is used as the feature vector of each tweet. In the second case, the [CLS] output from several layers are then combined with various methods to obtain the best possible feature to improve the recognition result at this stage. This procedure is carried out in both base and large modes with RoBERTa and BERT. Tables 4 and 5 show the results of these experiments.

*Table 4: Accuracy of emotion recognition using different combinations of [CLS] output layers of RoBERTa transformer (selecting appropriate features based on the whole sentence)*

| **Transformer** \ **Dataset** | | **Emotion Intensity(EI)** | **Implicit Emotion Shared Task (IEST)** | **EmoTex** |
|---|---|---|---|---|
| **RoBERTa base** | - | 90.04 | 67.03 | 89.16 |
| **RoBERTa large** | - | 91.1 | 68.07 | 90.3 |
| **RoBERTa large 24 layers** | Concatenate | **91.23** | **68.73** | **90.42** |
| | Average | 91.11 | 68.54 | 90.15 |
| | Sum | 91.18 | 68.7 | 90.37 |
| **RoBERTa large 4 last layers** | Concatenate | 91.21 | **68.73** | 90.4 |
| | Average | 91.15 | 68.66 | 90.38 |
| | Sum | **91.2** | **68.72** | **90.42** |
| **RoBERTa large 2 last layers** | Concatenate | 91.18 | 68.64 | 90.33 |
| | Average | 91.15 | 68.5 | 90.16 |
| | Sum | 91.16 | 68.61 | 90.35 |

*Table 5: Accuracy of emotion recognition using different combinations of [CLS] output layers of BERT transformer (selecting appropriate features based on the whole sentence)*

| **Transforme** \ **Dataset** | | **Emotion Intensity(EI)** | **Implicit Emotion Shared Task (IEST)** | **EmoTex** |
|---|---|---|---|---|
| **Bert base** | - | 88.36 | 65.02 | 87 |
| **Bert large** | - | 89.7 | 66.45 | 88.9 |
| **Bert large 24 layers** | Concatenate | **89.78** | **66.64** | 89.01 |
| | Average | 88.64 | 65.8 | 88.69 |
| | Sum | 89.5 | 66.51 | 88.67 |
| **Bert large 4 last layers** | Concatenate | 89.77 | 66.63 | 89 |
| | Average | 89.74 | 66.58 | 88.95 |
| | Sum | 89.76 | 66.63 | **89.02** |
| **Bert large 2 last layers** | Concatenate | 89.72 | 66.59 | 88.84 |
| | Average | 89.66 | 66.43 | 88.62 |
| | Sum | 89.73 | 66.58 | 88.85 |

The results from Tables 4 and 5 reveal that the RoBERTa outperforms BERT in extracting sentence features. It has resulted in more accurate emotion recognition. In each tables, the best performances are shown in bold. The analysis demonstrates that when it comes to extracting the features of the entire sentence, the best option is to "Concatenate the 24 layers", and with a very small distance, the "Sum of the last 4 layers", which do not differ much in performance.

### 4.2.2. Results of Feature Extraction in Words

In Section 3.3.2, all words' feature vectors are extracted with the purpose of paying more attention to words. The feature vector for each word is initially extracted from the last layer of transformer. The sentence feature vector is then created by merging these vectors. In the second step, the vectors of each word are extracted from several transformer layers and combined in various ways. Each word's vector is created by using these combinations. After the vectors of all words are merged, then the emotion recognition task is completed. The accuracy of emotion recognition is demonstrated in Tables 6 and 7.

*Table 6: Accuracy of emotion recognition using different combinations of RoBERTa transformer layers (selecting the appropriate combination of word feature vectors)*

| Transformer \ Dataset | | Emotion Intensity(EI) | Implicit Emotion Shared Task (IEST) | EmoTex |
|---|---|---|---|---|
| **RoBERTa base** | – | 91.79 | 69.22 | 90.52 |
| **RoBERTa large** | – | 92.53 | 69.4 | 91.17 |
| **RoBERTa large 24 layers** | Concatenate | **93.5** | **70.11** | **92.14** |
| | Average | 92.92 | 69.98 | 91.96 |
| | Sum | 92.9 | 69.95 | 91.85 |
| **RoBERTa large 4 last layers** | Concatenate | 93.47 | 69.96 | 92.12 |
| | Average | 93.35 | 69.87 | 92.08 |
| | Sum | **93.48** | **70.10** | **92.11** |
| **RoBERTa large 2 last layers** | Concatenate | 93.46 | 69.82 | 92.06 |
| | Average | 93.36 | 69.09 | 91.87 |
| | Sum | 93.47 | 69.87 | 91.94 |

The comparison of Tables 4 with 6 and 5 with 7 reveals that integrating all words' feature vectors improved recognition accuracy when compared to using [CLS] as the sentence's feature vector. The "summing the last 4 layers" method has the advantage of being flexible enough to reduce the dimensions without losing much more information as well as it is easy to use. It also takes less time to execute.

| *Table 7: Accuracy of emotion recognition using different combinations of BERT transformer layers (selecting the appropriate combination of word feature vectors)* | | | | |
|---|---|---|---|---|
| **Transformer** | | **Emotion Intensity(EI)** | **Implicit Emotion Shared Task (IEST)** | **EmoTex** |
| **Bert base** | - | 89.43 | 67.48 | 88.56 |
| **Bert large** | - | 90.71 | 67.92 | 90.38 |
| **Bert large 24 layers** | Concatenate | **90.98** | **68.03** | **90.6** |
| | Average | 90.73 | 67.76 | 89.88 |
| | Sum | 90.74 | 67.92 | 90.4 |
| **Bert large 4 last layers** | Concatenate | 90.93 | 68 | 90.54 |
| | Average | 90.9 | 67.94 | 90.47 |
| | Sum | 90.94 | 68 | 90.52 |
| **Bert large 2 last layers** | Concatenate | 90.86 | 67.98 | 90.4 |
| | Average | 90.8 | 67.90 | 90.45 |
| | Sum | 90.81 | 67.9 | 90.52 |

The results of Tables 4–7 show that RoBERTa outperforms BERT. Therefore, in CEFER, we use the RoBERTa transformer and the summing of the last four layers method for features extraction in words and feature and sentence feature vectors are constructed by merging word vectors.

#### 4.2.3. Results of CEFER using Meta-Embedding Vectors

According to meta-embedding proposed in Section 3.5, the final feature vector for each tweet is created. Then tweets are classified and assigned an emotional label. The results are shown in Table 8.

*Table 8: Improving the accuracy of emotion recognition using the proposed Meta-embedding*

| **Dataset** | **Bert Large (4 last layer)** | **Bert Large (4 last layer) +EmoSyn** | **RoBERTa Large (4 last layer)** | **RoBERTa Large (4 last layer) +EmoSyn** |
|---|---|---|---|---|
| **Emotion Intensity(EI)[40]** | 90.94 | 92.61 | 93.48 | **95.16** |
| **Implicit Emotion Shared Task (IEST)[41]** | 68 | 69.1 | 70.1 | **70.57** |
| **EmoTex[16]** | 90.52 | 92.07 | 92.11 | **94.93** |

Table 8 shows that the CEFER (the last column of the table) outperforms the RoBERTa and BERT. These findings support the need to reinforce the inference which mentioned in Section 3.4.

### 4.3. Comparing with other works

In this section, the results of comparison of CEFER with other similar studies are presented. The comparison is done from two perspectives: recognition of implicit emotions and recognition of explicit emotions which are presented in Section 4.3.1 and 4.3.2 respectively.

#### 4.3.1. Implicit Emotion Recognition

WASSA2018IEST [41], as explained in Section 4.1, is designed by masking all explicit emotional words in the text to detect the emotion of the sentence with the help of other words that express emotion implicitly. CEFER is evaluated on this dataset and compared to some similar studies. Results are shown in Table 9.

Thirty teams competed in the WASSA2018 competition, the majority of which used deep learning methods [20]. One of the greatest results was related to IIIDYT [42], which employed the Elmo and the BiLSTM network for classification and had F1-score of 71.05.

HUMIR [44] introduced a deep neural network model that used two BiLSTM networks to display the target word's left and right context in the same competition. The outputs were referred to as text-sensitive features. The authors also included vocabulary and emotion-based features into the BiLSTM model. With F1-score of 68.8, this model was ranked sixth among the 30 teams that competed.

*Table 9: Comparison of CEFER results with other methods of recognizing implicit emotions*

| Proposed Work | Dataset | Contribution | Best prediction |
|---|---|---|---|
| **IIIDYT [42]** | WASSA 2018 IEST[41] | Pre-trained ELMo + BiLSTM | 71.05 |
| **NTUA-SLP [43]** | | Ensemble of transfer learning techniques + LSTM | 70.9 |
| **CEFER** | | Combining RoBERTa +Emotional vector | 70.57 |
| **DuFER [15]** | | 2 kinds of feature sets, lexicon-based, 3language models weighting + SVM, Decision tree | 68.93 |
| **HUMIR [44]** | | 3 kinds of feature sets, lexicon-based, emotion-weight and context-sensitive BiLSTM + MLP | 68.8 |
| **SINAI[56]** | | four deep learning systems based on a sequence encoding layer + external emotional knowledge | 58.3 |

SINAI [56], which ranked 22nd out of 30 teams in the WASSA competition with a F1-score of 58.3 [20], provided four deep learning algorithms based on a sequence encoding layer. Systems that used external emotional knowledge had a higher generalization capacity according to their findings.

The fourth line of Table 9 shows DuFER [15], which is the prior framework established by the authors of this study. DuFER used emotional dictionaries as well as a feature selection and special feature weighting method. It has 68.93 F1-score. Based on the accuracy of other models in WASSA competition [20], DuFER is in sixth place. The F1-score of CEFER is 70.57, placing it third.

### 4.3.2 Explicit Emotion Recognition

Two datasets, EmoTex and Emotion Intensity are considered when examining CEFER outcomes on datasets with explicit emotions.

#### 4.3.2.1- Evaluation results of CEFER on EmoTex-dataset

A point should be mentioned before the results are presented. The Plutchik emotional model is used to classify emotions in CEFER. Happiness vs. sadness, anger vs. fear, trust vs. disgust, and surprise vs. expectation are among the eight bipolar emotions in this model. The Circumplex emotional model is used by the Emotex-method which considers four main classes: happy-active, happy-inactive, unhappy-active, and unhappy-inactive. The equivalence of these emotional classes are reported in Table 10 which is used to compare the results of the proposed framework and the Emotex-method on the EmoTex dataset. In this dataset, three frameworks are tested, as shown in Table 11.

1. To determine emotional classes, the Emotex-method employs four features. According to the results of classification based on these features in [16], the use of three features other than unigram has little influence on improving the results by 0.3 percent.
2. DuFER [15] recognizes emotions in tweets using a machine learning algorithm. By applying modified TF-IDF weights to words and maximum probability for expressions, it employs language models and computational linguistics. Finally, a combination classification method is used to determine the emotion of each tweet.
3. In CEFER, RoBERTa Large is used to extract the feature vector of each word. The emotional vector of the words is then retrieved and added to their feature vector using EmoSyn [15]. The acquired feature vectors are used to recognize the sentence emotion. Results are shown in the first to third rows of Table 11.

*Table 10: Equalizing emotional classes in Circumplex model and Emotional Classes [15]*

| **EmoTex Emotional Classes** | **DuFER Emotional Classes** |
|---|---|
| Unhappy Active | Fear, Disgust, Anger |
| Unhappy Inactive | Sadness, Pessimism |
| Happy Active | Surprise, Joy, love |
| Happy Inactive | Trust, Optimism, Anticipation |

#### 4.3.2.2- Evaluation results on the Emotion Intensity Dataset (EID)

A tweet, its emotional label, and a value associated with the intensity of emotion are all included in the EID dataset [40]. There are four different emotions in this dataset. The purpose of this work is to discover the emotional label for each tweet without using emotional intensity values. As a result, CEFER only uses tweets and emotional labels that are related to them. The $4^{th}$ till $6^{th}$ rows of Table 11 illustrate the outcomes of the CEFER framework and two other methods. The findings obtained from this dataset are superior to the results obtained from other datasets due to the repetition of different emotional terms in a tweet and the existence of more emotional words in the dataset.

In [45], a model based on LSTM and CNN are proposed which using the level of attention. The achievement of 94% accuracy for the model shows the effectiveness of the approach although the model has high complexity. In DuFER [15], the accuracy 95.27 for explicit emotions is achieved and the result for CEFER is 95.16.

*Table 11: Comparison of CEFER with other explicit emotion recognition methods*

| Proposed Work | Dataset | Contribution | Best prediction |
|---|---|---|---|
| **Emotex-method [16]** | EmoTex-dataset[16] | Unigram, emoticons, punctuation, and negation as features + Naïve Bayes, SVM | 90 |
| **CEFER** | | RoBERTa $_{Large}$ (Sum 4 last layer)+EmoSyn | 94.93 |
| **DuFER[15]** | | 2 kinds of feature sets, lexicon-based, 3language models weighting + SVM, Decision tree | 95.12 |
| **Method in [45]** | EID from Semeval2018[40] | Deep neural networks, Bi-LSTM, CNN+ FastText word-embedding | 94 |
| **CEFER** | | RoBERTa $_{Large}$ (Sum 4 last layer)+EmoSyn | 95.16 |
| **DuFER[15]** | | 2 kinds of feature sets, lexicon-based, 3language models weighting + SVM, Decision tree | 95.27 |

## 5. Discussion

According to the results are shown in Tables 9 and 11, the proposed framework CEFER scored better on implicit recognition than explicit since in constructing the feature vector of each word, the transformer dynamically uses the words around it and also the meaning of the whole sentence. Therefore, the vector of each word implicitly contains the entire sentence's meaning. On the other hand, the use of the MASK feature in the transformer corresponds to the masking nature of some words (words that explicitly express emotion) in the IEST database. Each text in this dataset implicitly expresses an emotion. Hence, it can be said that in CEFER, the appropriateness of the method for extracting feature vectors is the most important reason for improving the recognition accuracy in the IEST dataset. In fact, it can be seen that CEFER is able to use and deduce the text well even without any explicit emotional words.

However, despite the high accuracy achieved by CEFER, it performs worse in explicit emotions than DuFER, the authors' earlier study. We return to the structure of the two methods in order to figure out what is causing this issue. The use of the modified TF-IDF in DuFER, which is based on the BOW technique, causes all emotional words related to one emotion to fall into the same class. As a result, the weight of that emotion increases. CEFER employs a transformer that encodes each word according to its context. So, the entire sentence contributes to the creation of the word feature vector.

As a result, the feature vector of a specific emotive term does not remain consistent across the dataset and changes from sentence to sentence. The chances that the term will fall into the appropriate emotional category is low. Third column of Table 8 before employing the emotional vector demonstrates this. The usage of EmoSyn in creating an emotional vector and then adding it to the feature vector of words improved the accuracy of emotion recognition in each tweet in the proposed framework. The authors believe that concentrating on the classification module will aid future studies in improving recognition accuracy. Tables 12 and 13 show the characteristics of the reviewed papers as well as the proposed framework. Table 12 examines the quality of articles in terms of key parameters and Table 13 studies information sources employed by different papers.

*Table12: Qualitative analysis of the relevant literature according to key parameters. I and E indicate implicit and explicit respectively*

| Paper | Embedding | Transformer | Dataset | Algorithm | Ensemble | Evaluation Measure | Implicit/ Explicit | Emotions detected |
|---|---|---|---|---|---|---|---|---|
| **15 Khoshnam 2022** | — | — | EmoTex, Emotion Intensity, SemEval2018(E-c) | SVM, Decision tree | ✓ | 95.12 (ACC) | I/E | 8 Plutchik emotions |
| | | | Implicit Emotion Shared Task (IEST) | | | 68.93 (F1) | | |
| **23 Park 2019** | — | — | Self-created | SVM, Logistic Regression, Neural Network | — | 61 (ACC) | E | Joy, Sad, Angry, Fear |
| **24 Chatterjee 2019** | Glove, SSWE | — | Self-created From Twitter | LSTM | — | 73.55 (F1) | E | Joy, Sad, Angry |
| **25 Al-Omari 2020** | Glove | BERT | SemEval2019 | BiLSTM, RNN | ✓ | 74.78 (F1) 91.9 (ACC) | E | Joy, Sad, Angry, Other |
| **26 Illendula 2019** | FastText | — | Self-created | BiLSTM , Attention | ✓ | 72 (ACC) | E | Joy, Sad, Angry, Fear, Love, Surprise, Thankful |
| **27 Polignano 2019** | FastText | — | SemEval2018, SemEval2019 | BiLSTM, CNN, Attention | ✓ | 84 (ACC) | E | Anger, Fear, Joy, Sad |
| **28 Zahiri 2018** | Word2vec | — | Self-created on Friends TV Show | CNN, Attention | — | 54 (ACC) | E | Joy, Sad, Peaceful, Powerful, Scared, Mad, Neutral |
| **29 Jabreel 2019** | — | — | SemEval2018 (Task1) multi-label | BiRNN, Attention | — | 59 (Jaccard ACC) | E | 8 Plutchik emotions |
| | | | | | | 70.1 (F1) | | |
| **30 Cortiz 2021** | ___ | Bert, Distil BERT, RoBERTa, XLNet | ISEAR, SemEval2018(Task 1) SemEval2019(Task 3) | BiLSTM, CNN | ✓ | 74.2 (ACC) 49 (F1) | E | Joy, Sad, Angry, Other |
| **31 Schmidt 2021** | ___ | GBert-large | Self-created on 11 German historical plays | Transformer, historical knowledge | ✓ | 66 (ACC) 90 (ACC) | E | 6 Main 13 Sub emotion |
| **42 Balazs 2018** | ELMO | ___ | Implicit Emotion Shared Task (IEST) | BiLSTM | ✓ | 71 (F1) | I | 6 of 8 Plutchik emotions ( sad, joy, disgust, surprise, anger, fear) |
| **43 Chronopoulou 2018** | Word2vec | ___ | Implicit Emotion Shared Task (IEST), Semeval-2017(Task4) | LSTM, Self-attention, Transfer-learning | ✓ | 70.3 (F1) | I | 6 of 8 Plutchik emotions ( sad, joy, disgust, surprise, anger, fear) |
| **44 Naderalvojoud 2018** | Glove | ___ | Implicit Emotion Shared Task (IEST) | BiLSTM | ✓ | 68.6 (F1) | I | 6 of 8 Plutchik emotions ( sad, joy, disgust, surprise, anger, fear) |
| **52 Krommyda 2021** | ___ | ___ | Self-created on Social media posts | A hybrid rule-based, LSTM Network | ✓ | 91.9 (ACC) | E | 8 Plutchik emotions |
| **53 Choudrie 2021** | ___ | RoBERTa | Self-created on CrowdFlower dataset and the selected Reddit posts | LSTM, BiLSTM , Transfer learning | ✓ | 80.33 (ACC) 75.25 (F1) | E | anger, boredom, empty, enthusiasm, fun, happiness, hate, love, neutral, relief, sadness, surprise and worry |
| **56 Plaza-del-Arco 2018** | Glove | ___ | Implicit Emotion Shared Task (IEST ) | LSTM | ___ | 58.3 (F1) | I | 6 of 8 Plutchik emotions ( sad, joy, disgust, surprise, anger, fear) |
| **CEFER Proposed 2022** | ___ | RoBERTa, BERT | EmoTex, Emotion Intensity, SemEval2018(E-c)<br>Implicit Emotion Shared Task (IEST), | Transformer, Meta embedding, FF Neural Network | ✓ | 95.16 (ACC)<br>70.57 (F1) | I/E | 8 Plutchik emotions |

*Table13: Overview of information sources employed by different papers*

| Paper | Word | Sentence | Document | Emoji | Lexicon | Emotional embedding |
|---|---|---|---|---|---|---|
| 15 Khoshnam(2022) | ✓ | | | | ✓ | |
| 23 Park(2019) | ✓ | ✓ | | ✓ | | |
| 24 Chatterjee(2019) | ✓ | | | | | ✓ |
| 25 Al-Omari(2020) | ✓ | | ✓ | ✓ | ✓ | |
| 26 Illendula(2019) | ✓ | | | ✓ | | |
| 27 Polignano(2019) | ✓ | | | | | |
| 28 Zahiri(2018) | ✓ | | | | | |
| 29 Jabreel(2019) | ✓ | | | | | ✓ |
| 30 Cortiz(2021) | | ✓ | | | | |
| 31 Schmidt(2021) | | | ✓ | | | ✓ |
| 42 Balazs(2018) | ✓ | ✓ | | ✓ | | |
| 43 Chronopoulou(2018) | ✓ | ✓ | | | | |
| 44 Naderalvojoud(2018) | | ✓ | | | ✓ | |
| 52 Krommyda(2021) | ✓ | | | ✓ | ✓ | |
| 53 Choudrie(2021) | ✓ | | | | | |
| 56 Plaza-del-Arco(2018) | ✓ | | | ✓ | ✓ | |
| CEFER (2022) | ✓ | ✓ | | | ✓ | ✓ |

## 6. Conclusions

In this Paper, we provide a four facets framework for recognizing emotions in short texts. We developed and evaluated a supervised deep learning framework called CEFER, based on four important aspects of text processing and emotion recognition. CEFER automatically classifies short texts with implicit and / or explicit emotions. Our experiments show that CEFER correctly classifies emotions in nearly 95% of text messages with explicit emotions. It also achieves 70.57% in classifying texts with implicit emotions, which is a very good result compared to the methods proposed in the IEST competition and it is in third place (Table 9). CEFER focuses on text processing at both word and sentence levels. It acquires textual properties using context properties by Roberta. Then, according to the emotional features of each word, it builds an emotional vector using an emotional embedding. Combining these features in meta embedding deals with the emotional recognition of tweets. We show that using words and sentence features simultaneously has a higher performance than using them individually and improves recognition results compared to transformers. We also show that paying attention to implicit emotions also improves the power of recognizing explicit emotions. Our future plan is concentrating on the classification module to improve recognition accuracy.